\documentclass[lettersize,journal]{IEEEtran}
\pdfoutput=1
\usepackage{amsmath,amsfonts}
\usepackage{algorithmic}
\usepackage{algorithm}
\usepackage{array}
\usepackage[caption=false,font=normalsize,labelfont=sf,textfont=sf]{subfig}
\usepackage{textcomp}
\usepackage{stfloats}
\usepackage{url}
\usepackage{verbatim}
\usepackage{graphicx}
\usepackage{cite}

\usepackage{longtable} 
\usepackage{multirow}
\usepackage{subfig}
\usepackage{booktabs}

\begin{document}

\title{MCformer: Multivariate Time Series Forecasting with Mixed-Channels Transformer}

\author{Wenyong Han, Tao Zhu~\IEEEmembership{Member,~IEEE}, Liming Chen~\IEEEmembership{Senior Member,~IEEE} \\ 
Huansheng Ning~\IEEEmembership{Senior Member,~IEEE}, Yang Luo~\IEEEmembership{Member,~IEEE}, Yaping Wan~\IEEEmembership{Member,~IEEE}


\thanks{This work was supported in part by the National Natural Science Foundation of China(62006110), the Natural Science Foundation of Hunan Province (2021JJ30574). (Corresponding author: Tao Zhu.)}
\thanks{Wenyong Han, Yang Luo, Tao Zhu and Yaping Wan are with the School of Computer Science, University of South China, 421001 China (e-mail:wyhan@stu.usc.edu.com, luoyang@usc.edu.cn,  
tzhu@usc.edu.cn, ypwan@aliyun.com).}
\thanks{Liming Chen was with School of Computer Science and Technology, Dalian University of Technology, China. Huansheng Ning was Department of Computer \& Communication Engineering, University of Science and Technology Beijing, 100083 China (email: l.chen@ulster.ac.uk, ninghuan-sheng@ustb.edu.cn).}
}


\IEEEpubidadjcol

\maketitle

\begin{abstract}
The massive generation of time-series data by large-scale Internet of Things (IoT) devices necessitates the exploration of more effective models for multivariate time-series forecasting. In previous models, there was a predominant use of the Channel Dependence (CD) strategy (where each channel represents a univariate sequence). Current state-of-the-art (SOTA) models primarily rely on the Channel Independence (CI) strategy. The CI strategy treats all channels as a single channel, expanding the dataset to improve generalization performance and avoiding inter-channel correlation that disrupts long-term features. However, the CI strategy faces the challenge of inter-channel correlation forgetting. To address this issue, we propose an innovative Mixed Channels strategy, combining the data expansion advantages of the CI strategy with the ability to counteract inter-channel correlation forgetting. Based on this strategy, we introduce MCformer, a multivariate time-series forecasting model with mixed channel features. The model blends a specific number of channels, leveraging an attention mechanism to effectively capture inter-channel correlation information when modeling long-term features. Experimental results demonstrate that the Mixed Channels strategy outperforms pure CI strategy in multivariate time-series forecasting tasks.
\end{abstract}


\section{Introduction}

\IEEEPARstart{W}{ith} the widespread application of Internet of Things (IoT) devices in fields such as meteorology\cite{angryk2020multivariate,han2021joint,cheng2018deepiot}, traffic\cite{ghosh2009multivariate,cirstea2022traffic2,qin2023traffic3,zhou2020reinforced}, and electricity\cite{stefenon2023wavelet}, the increasing number of devices has resulted in the generation of a significant amount of time-series data. These data can be utilized for decision-making\cite{han2021multivariateiot}, resource allocation\cite{hua2023kae}, and forecasting future trends\cite{yi2023frequency,ni2023basisformer}, thereby enhancing the efficiency and reliability of IoT systems. Time-series forecasting tasks arising from IoT devices aim to forecast future states based on historical data. Given that IoT data typically exhibits characteristics such as non-linearity, rapid sampling, and multi-channel aspects, this task poses certain challenges.

\begin{figure}[!t]
    \centering
    \includegraphics[width=1\linewidth]{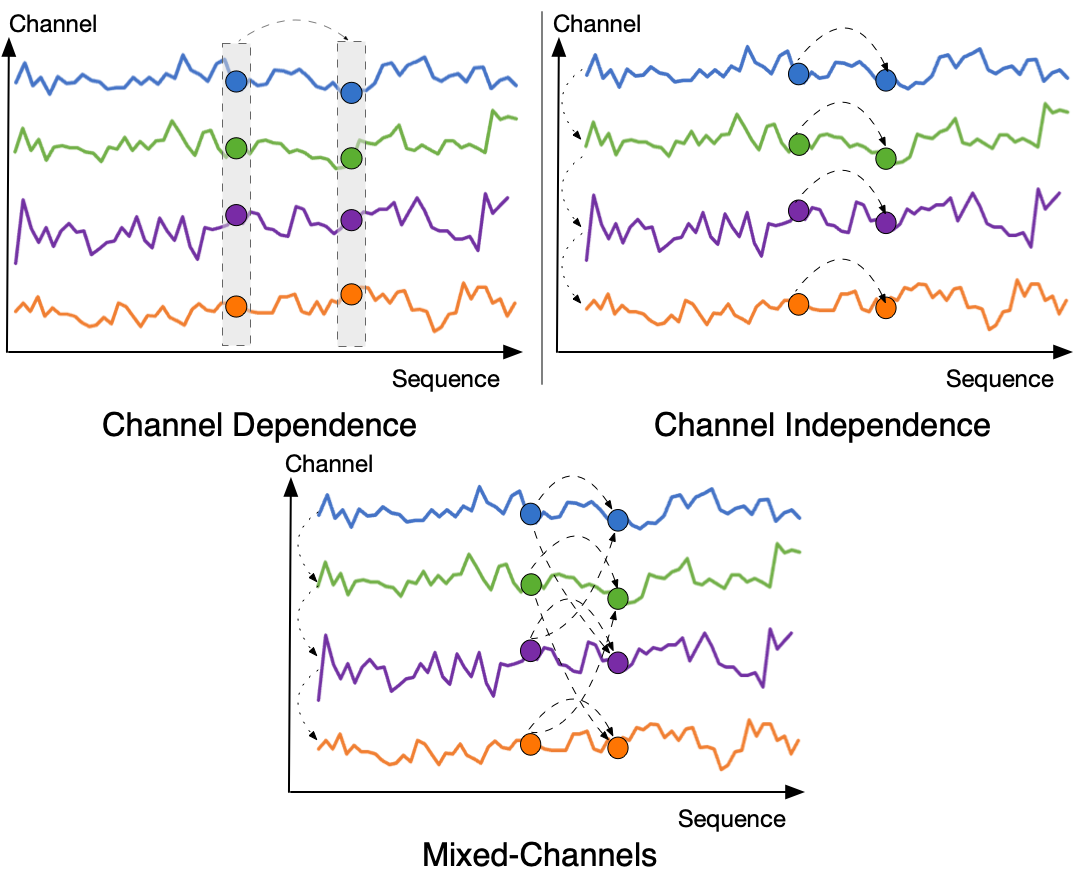}
    \caption{The difference between the CI, CD and Mixed Channels strategies.}
    \label{differ m and u}
\end{figure}

Due to the typically longer sampling intervals and numerous sampling channels in time-series data generated by IoT devices, when dealing with multivariate time series from such IoT devices, it is necessary to consider both long sequence modeling and the complex interrelationships between multiple channels. Given the outstanding capability of Transformers in modeling long sequences demonstrated in the field of natural language processing, this ability is also crucial in time-series forecasting tasks, leading to the emergence of models like LogTrans\cite{li2019logtrans},  Informer\cite{zhou2021informer}, Reformer\cite{kitaev2020reformer}, Autoformer\cite{wu2021autoformer}, FEDformer\cite{zhou2022fedformer}, ScaleFormer\cite{shabani2022scaleformer}, Pyraformer\cite{liu2022pyraformer}, FPPformer
\cite{chen2021fpp}, and others. These models have made significant progress in the realm of long time-series modeling.In recent years, some research has shifted focus to the challenges of multivariate time series, exemplified by models such as Crossformer\cite{zhang2023crossformer}  and SageFormer\cite{zhang2023sageformer}. These models undertake learning across all channels, with a specific focus on capturing dependencies between multiple variables. All these models can be considered as Channel Dependence (CD) Strategy models (A univariate sequence is treated as a channel),This approach takes multivariate data as a whole input and allows the model to learn the correlation between channels, as shown in the Fig. \ref{differ m and u}.

However, these CD strategy models also have drawbacks. DLinear\cite{zeng2023dlinear} has surpassed existing models with a simple architecture. PatchTST\cite{Yuqietal-2023-PatchTST} introduces a Channel Independent (CI) strategy model, further improving the state-of-the-art (SOTA). The CI strategy treats all channels as a single channel, thereby expanding the dataset and enhancing the model's generalization capability, as shown in the Fig. \ref{differ m and u}. The success of the CI strategy has drawn attention to the impact of both CD and CI strategies on models, as seen in models like PRReg\cite{han2023capacity}, PETformer\cite{lin2023petformer}, CSformer\cite{wang2023csformer}, itransformer\cite{liu2023itransformer}, and others. Subsequently, TiDE\cite{das2023tide} (a CI model based on MLP) not only performs similarly to PatchTST but also excels in spatiotemporal efficiency. Research on PETformer found that channel independence is superior to channel dependence, possibly because multivariate features can interfere with the extraction of long sequence features. This result goes against intuition, as in deep learning, more information typically improves model generalization.

In summary, there are two main reasons why existing SOTA models are mostly based on the CI strategy: firstly, the CI strategy can expand the dataset to improve the generalization performance of the model, as seen in PatchTST; secondly, the CI strategy can avoid the destruction of long-term feature information by channel-wise correlation information, as demonstrated by  PETformer. However, the CI strategy also has drawbacks, as it tends to overlook inter-channel feature information. In cases with a large number of channels, there may be an issue of inter-channel correlation forgetting, akin to the forgetting of long-term information in RNNs\cite{lipton2015critical}. In this context, we propose a Mixed Channels strategy. This strategy retains the advantages of the CI strategy in expanding the dataset while effectively avoiding the disruption of long-term feature information by channels. It also addresses the issue of inter-channel correlation forgetting. Fig. \ref{differ m and u} illustrates the differences between the Mixed Channels strategy and CI and CD. Based on the aforementioned strategy, we propose a multi-channel time series forecasting model with mixed channel features. Fig. \ref{MC_method} provides an overview of the model. Specifically, our model first expands the data using the CI strategy, then mixes a specific number of channels, and allows the attention mechanism to effectively capture the correlation information between channels when modeling long-term feature information. Finally, the encoder result is unflattened to obtain the predicted values of all channels. The contributions of our proposed model can be summarized as follows:
\begin{figure*}[!t]
    \centering
    \includegraphics[width=1\linewidth]{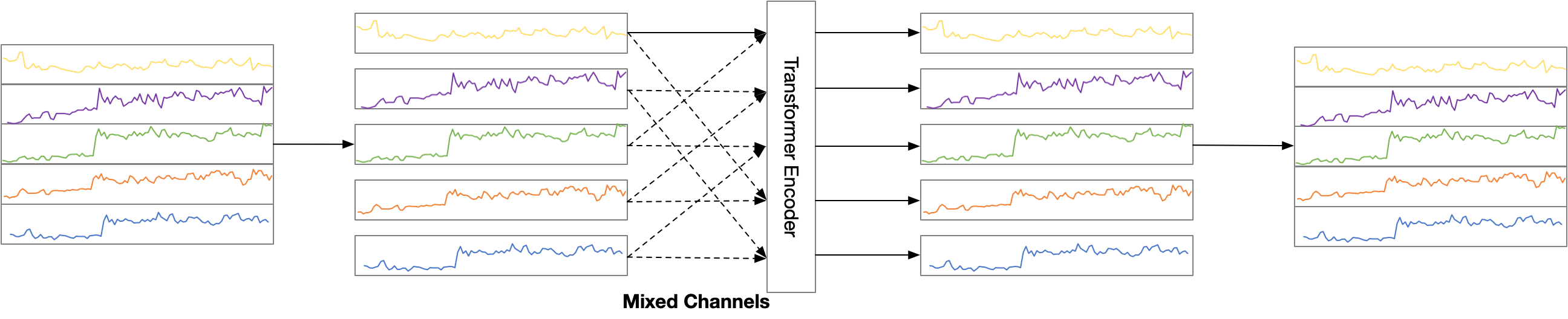}
    \caption{
Overview of the Mixed Channels method: Multivariate time series data is initially decomposed by channel, resulting in individual channel data. Subsequently, based on the channel interval size, data from different channels is mixed. The mixed data will share parameters in the Transformer Encoder.}
    \label{MC_method}
\end{figure*}

\begin{itemize}
    \item We propose a Mixed Channels strategy that seeks to minimize the drawbacks of channel features disrupting long-term information under the CD strategy, while retaining the advantages of expanding the dataset under the CI strategy. This enables the model to more effectively learn inter-channel dependency information.

    \item Based on the Mixed Channels strategy, we present a Multivariate time-series forecasting model with mixed channel features. By employing the Mixed-Channels Block, the model expands the dataset and integrates inter-channel dependency information through a blended approach.

    \item Furthermore, MCformer has been experimentally evaluated on five real-world Multivariate datasets, achieving outstanding results compared to the current SOTA. We conduct two ablation experiments, investigating the impact of the Mixed-channels approach on datasets with different channel quantities and the influence of varying channel mixing quantities on model performance. This further substantiates the effectiveness of our approach. Additionally, we explore the rich feature relationships among multiple channels. Using correlation analysis, we present changing curves in the inter-channel correlations across multiple datasets, illustrating the dynamic evolution of feature relationships among channels over time.
\end{itemize}

\section{RElATED WORK}
\textbf{Long Time Series Forecasting.} In the past, long time series forecasting has been a focal point in the field of time series analysis. The approaches to address this problem can be broadly categorized into several models: statistical methods, MLP-based methods, CNN-based methods, RNN-based methods, and Attention-based methods. Statistical models like ARIMA\cite{bartholomew1971arima} assume time series stationarity, but many real-world time series are non-stationary. RNN-based models (such as LSTNet\cite{lai2018lstnet} and DeepAR\cite{salinas2020deepar}) excel in capturing sequence features, but their reliance on hidden states for feature propagation makes them notably deficient in modeling long sequences. In recent years, CNN-based models (like MICN \cite{wang2022micn}and TimesNet\cite{wu2023timesnet}) have emerged, leveraging the outstanding performance of CNNs in the image domain to model multivariate and long sequence features. With the advent of Transformers, which have demonstrated excellent performance in natural language processing, they have gained attention in the time series domain as well. Models like Informer\cite{zhou2021informer}, Autoformer\cite{wu2021autoformer}, and FEDformer\cite{zhou2022fedformer} have achieved SOTA by building upon and improving the Transformer. Informer effectively extracts critical information by employing an improved ProbSparse self-attention mechanism. Autoformer captures both local and global features of time series by leveraging the decomposition and autocorrelation concepts from traditional time series analysis. FEDformer, on the other hand, transforms attention from the time domain to the frequency domain, reducing complexity through Fourier transformation. MLP-based models, gaining more attention in long sequence forecasting after the publication of Dlinear\cite{zeng2023dlinear}, have spurred questioning of the modeling capability of Transformers in time series. Subsequently, PatchTST\cite{Yuqietal-2023-PatchTST}, utilizing the native ViT\cite{dosovitskiy2020vit} with a single-channel strategy, surpassed Dlinear once again.

\textbf{Multivariate Time series Forecasting.} In the realm of multivariate time series forecasting, the challenge of long time series forecasting has persistently garnered attention. However, in recent years, especially with the powerful complex feature extraction capabilities demonstrated by Transformers across various modalities of data, more research in multivariate time series forecasting has started focusing on a model's ability to model interactions among multiple variables. We briefly summarize some research on Multivariate feature modeling in previous models based on the CD strategy. For instance, models like TimesNet\cite{wu2023timesnet} and LSTnet \cite{lai2018lstnet} leverage CNNs to capture dependencies across dimensions. In Former-based models, such as Crossformer\cite{zhang2023crossformer}, the input multivariate time series is embedded into a 2D vector array to retain both temporal and dimensional information. A Two-Stage Attention (TSA) layer is then introduced to efficiently capture dependencies across time and dimensions.The Client\cite{gao2023client} and iTransformer\cite{liu2023itransformer} models differ from traditional enhanced Transformer models in that they transform the function of Attention from learning long-term features to learning inter-variable features. They utilize linear modules to learn trend information and attention modules to capture cross-variable dependency relationships.SageFormer\cite{zhang2023sageformer}is a Series-aware Graph-enhanced Transformer model designed to effectively capture and model dependencies between series using a graph structure. Although PatchTST\cite{Yuqietal-2023-PatchTST} employs a CI strategy for training, it still learns cross-variable features through model parameterization. The recent TiDE\cite{das2023tide}(MLP-Based Model) follows a similar strategy. Due to the impact of PatchTST, analyses in PRReg\cite{han2023capacity} and PETformer\cite{lin2023petformer} have explored the reasons for the lower effectiveness of CD strategies compared to CI strategies, proposing methods to enhance the effectiveness of CD strategies.

\section{METHOD}
\begin{figure*}
    \centering
    \includegraphics[width=1\linewidth]{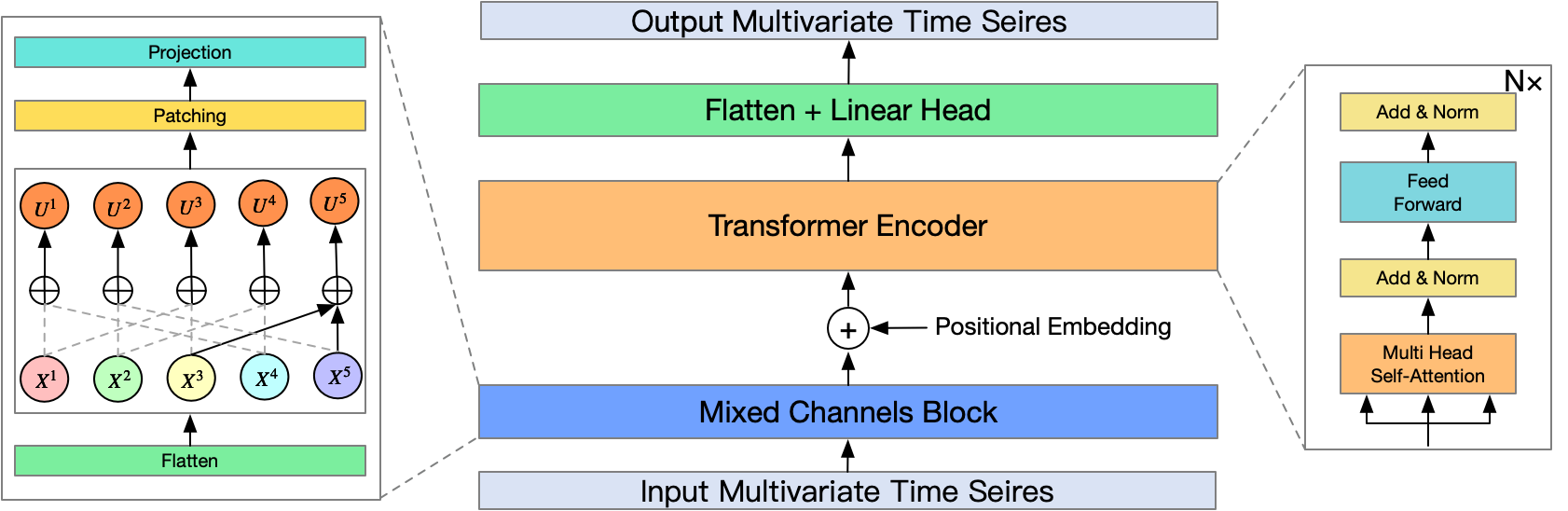}
    \caption{Mixed Channels architecture: In the Mixed-Channels Block, we decompose multivariate time series data into single channels and then blend the data from different channels. The blended data is then segmented into multiple patches, with each patch composed of adjacent samples. These patches are transformed into input tokens through a projection process.}
    \label{Mixed_channels_arch}
\end{figure*}

\subsection{Problem Definition}
In the task of multivariate time series forecasting, historical observations are represented as $X=\left \{\mathbf{x}_1, \mathbf{x}_2, ..., \mathbf{x}_t \right \} \in\mathbb{R}^{t\times M}$, where $t$ is the time step and $M$ is the number of variables. The observations at each time step $t$ are represented by a $M$-dimensional vector: $\mathbf{x}_t=[x_t^1, x_t^2, …, x_t^M] ^\top $. Our objective is to predict the multivariate observations at future time steps based on past observations. The problem can be formalized as follows: given the observed sequence $\{ \mathbf{x}_1, \mathbf{x}_2, ..., \mathbf{x}_{t}\}$ up to time step $t$, predict the multivariate observation sequence $\{ \mathbf{x}_{t+1}, ..., \mathbf{x}_{t+h} \}$ for time steps $t+1$ to $t+h$, where $h$ is the number of time steps to be predicted. We incorporate a Mixed-Channels Block into the vanilla Transformer Encoder to expand the dataset and blend inter-channel dependency information.
The architecture of our MCformer model is shown in Fig. \ref{Mixed_channels_arch}.

\subsection{Reversible instance Normalization}
The introduction of Reversible Instance Normalization (RevIN)\cite{kim2021revin} aims to address the issue of non-uniform temporal distribution between training and testing data, commonly referred to as distribution shift. Before the Mixed Channels module, we apply instance normalization to normalize each channel's data. A single channel is represented as \(\mathbf{x}^i=[x_1^i,x_2^i,…,x_t^i]\), where for each instance \(x^i_t\), we calculate the mean and standard deviation. After obtaining the forecasting results, these non-stationary information components are added back to the predicted values.

\begin{equation}
    \label{revin}
    RevIN(\mathbf{x}^i)=\left \{ \gamma _i\frac{\mathbf{x}^i-Mean(\mathbf{x}^i)}{\sqrt{Var(\mathbf{x}^i)+\varepsilon } }  \right \} ,i=1,2,\cdots ,M
\end{equation}

\subsection{Mixed-Channels Block}
We introduce a method called the Mixed Channels Module to enhance the representation of multivariate time series datasets.

\textbf{Flatten} We employ a Channel Independent (CI) strategy to flatten the data from $M$ channels. For a given sample $X$, after flattening, we obtain $X_F = \text{Flatten}(X) \in \mathbb{R}^{tM \times 1}$. The flattened $X_F$ is then treated as if it were $M$ individual samples.

\textbf{Mixed Channels}
 Mixed Channels involves combining data from different channels after Flatten. We perform the mixed channels operation through the following steps:
\begin{enumerate}
    \item Compute Interval Size: We calculate the interval size \(\left\lfloor\frac{M}{m}\right\rfloor\), where $m$ is the number of channels to be mixed.
    \item Mixed Channels Operation: For a given time step $t$, starting from the target channel, we stack every other channel at an interval stride to form $U^i\in \mathbb{R}^{t\times m}$. Specifically, the output of the Mixed Channels Module is defined as:\\
\end{enumerate}
\begin{equation}
\begin{aligned}
    \label{mixedchannels}
    U^i&=MixedChannels(\mathbf{x}^i,m)\\&=[stack(\mathbf{x}^i,C^1,C^2,…,C^m)]
\end{aligned}
\end{equation}

where $C^i$ represents the $i$-th channel taken at the $i$-th interval, and \(1 \leq i \leq m\).
By introducing the Mixed Channels Module, our aim is to enhance the expressive power of the input data, introducing multi-channel information to better capture the features of the time series data.

\textbf{Patch and Projection }In the current research \cite{Yuqietal-2023-PatchTST,lin2023petformer,das2023tide}, it has been observed that, compared to using time-point data as input, employing patches can better capture local information and also encompass richer dependencies between variables. 
Therefore, we utilize Patch to aggregate the sequence after mixing channels, and employ a single-layer MLP to project channel dependencies as well as adjacent temporal dependencies. This is expressed as:

\begin{equation}
\label{patch}
\mathcal{P}^i=Projection(Patch(U^i))
\end{equation}

\begin{figure}[H]
    \centering
    \includegraphics[width=1\linewidth]{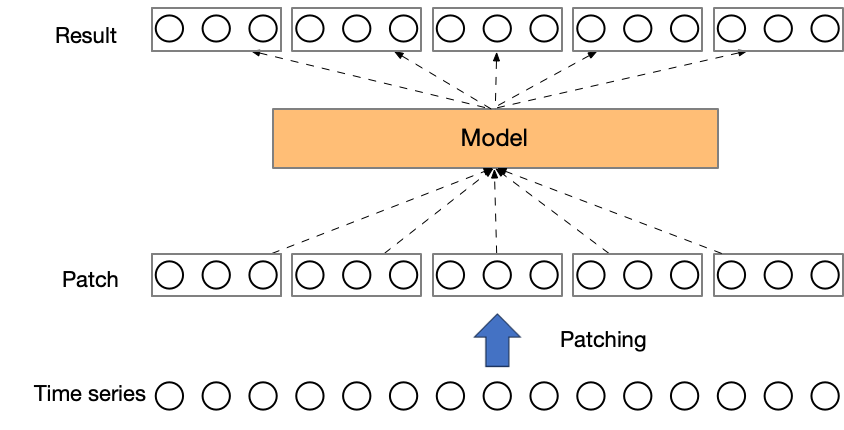}
    \caption{Application of Patch to temporal models. 
Using patches significantly extends the historical time range of the input while maintaining the same token length.}
    \label{patch fig}
\end{figure}

Here $\mathcal{P}^i\in \mathbb{R}^{P\times N} $, where $P$ is length after Projection, $N$ is the number of patches, \(N=\left \lfloor \frac{(L-p)}{S} \right \rfloor +2\), and $p$ is the length of Patch and $S$ is the stride length. This patch approach allows for the retention of both time dependencies and dependencies between multiple channels. It not only preserves the Transformer input's tokens but also further increases the size of the forecasting window while maintaining time dependencies. You can see in Fig. \ref{patch fig} how patches are utilized in time series tasks.
 
\begin{algorithm}[!t]
\caption{Pseudocode of MCformer Training}\label{alg:alg1}
\begin{algorithmic}
\STATE 
\STATE {\textbf{Input}} Historical traffic observations $X$ and future ground truth $Y_h$.
\STATE {\textbf{Output}} Predicted future observations $\hat{Y}_h$.
\STATE Shuffle data;
\STATE set $m \leftarrow stack\_len,p \leftarrow patch\_size$;
\STATE \textbf{for} each batch of  in data \textbf{do}
\STATE \hspace{0.5cm} \textbf{for} $i = 1\rightarrow M$ \textbf{do}
\STATE \hspace{1cm} The mean and variance is calculated by formula(\ref{revin});
\STATE \hspace{1cm} $\mathbf{x}^i_{norm} \leftarrow RevIN(\mathbf{x}^i)$, normalize each instance;
\STATE \hspace{0.5cm} \textbf{end}
\STATE \hspace{0.5cm} $X_F\in \mathbb{R}^{tM \times 1} \leftarrow$ flattening all the $\mathbf{x}^i_{norm}$;\\
\STATE \hspace{0.5cm} $U\in \mathbb{R}^{tM\times m} \leftarrow $ calculate by formula (\ref{mixedchannels});
\STATE \hspace{0.5cm} $\mathcal{P}\in \mathbb{R}^{M \times P\times N} \leftarrow $ the patches are divided and Projec-\\  tion by formula(\ref{patch});
\STATE \hspace{0.5cm} $\hat{Y}_h\in \mathbb{R}^{hM \times 1} \leftarrow $ modeling with encoder;
\STATE \hspace{0.5cm} $\hat{Y}_h\in \mathbb{R}^{h \times M} \leftarrow $ Unflatten by forecasting window length \\ $h$;
\STATE \hspace{0.5cm} \textbf{for} $t = 1 \rightarrow h$ \textbf{do}
\STATE \hspace{1cm} $\hat{Y}_{denorm} \leftarrow $denormalize $\hat{Y}_h$ each instance;
\STATE \hspace{0.5cm} \textbf{end}
\STATE \hspace{0.5cm} $loss \leftarrow$ calculate by formula(\ref{MSE});
\STATE \hspace{0.5cm} $Optimize(loss)$;
\STATE \textbf{end}
\end{algorithmic}
\label{alg1}
\end{algorithm}

\subsection{Encoder}
We employ the native Transformer encoder\cite{vaswani2017attention,dosovitskiy2020vit} to model the long-term and cross-variable features of the sequence, akin to the approach taken in PatchTST\cite{Yuqietal-2023-PatchTST}. As the Transformer does not explicitly model the sequence's order, to provide positional information, we utilize learnable additive positional encoding 
\(\mathcal{W}_{pos}\in \mathbb{R} ^{P\times N}\). The positional encoding is added to the embedded representation of the input sequence \(\mathcal{X}_{in}^i=\mathcal{P}^i+\mathcal{W}_{pos}\). This way, the model can differentiate elements at different positions.

The encoder consists of multiple layers with the same structure, and each layer comprises two sub-layers: the Multi-Head Self-Attention Layer and the Feedforward Neural Network Layer. The Multi-Head Self-Attention Layer is the first sub-layer of the encoder. In this layer, each element in the input sequence can interact with every other element, not just its neighboring elements. Three learnable matrices \(\mathcal{W}^Q\), \(\mathcal{W}^K\), and \(\mathcal{W}^V\) are used to compute \(Q^i=(\mathcal{X}_{in}^i)^T\mathcal{W}^Q\),\(K^i=(\mathcal{X}_{in}^i)^T\mathcal{W}^K\), and \(V^i=(\mathcal{X}_{in}^i)^T\mathcal{W}^V\). This is achieved by calculating attention weights, where each element receives a set of weights indicating its importance to other elements. With the use of a multi-head mechanism, the model can learn different aspects of attention, enabling it to better capture information within the sequence:

\begin{equation}
\label{attention}
    Attention(Q^i,K^i,V^i) = Softmax(\frac{Q^i(K^i)^T}{\sqrt{d_k} } )V^i
\end{equation}

The Feedforward Neural Network Layer comes after the Multi-Head Self-Attention Layer. The representation at each position is further processed through a fully connected feedforward neural network. This network typically consists of two fully connected layers, and its output is added to the input, creating a residual connection. This helps alleviate the gradient vanishing problem during training.

Residual connections and Layer Normalization are employed between the input and output of each sub-layer. This inclusion of residual connections makes it easier for the model to learn identity mappings, facilitating the training of deep networks. Additionally, the output of each sub-layer undergoes layer normalization to stabilize the training process.

\subsection{Loss Function}
We chose MSE (Mean Squared Error) and MAE (Mean Absolute Error) losses to evaluate the disparity between the model's forecastings and the actual values. MSE measures the model performance by calculating the average of the squared differences between the predicted and actual values:
\begin{equation}
    \label{MSE}
    \mathrm{MSE}=\frac1h\sum_{i=1}^h(y_i-\hat{y}_i)^2
\end{equation}

MAE assesses the model performance by computing the average of the absolute differences between the predicted and actual values:
\begin{equation}
    \label{MAE}
    \mathrm{MAE}=\frac1h\sum_{i=1}^h|y_i-\hat{y}_i|
\end{equation}

We have introduced the main structure of MCformer. Algorithm \ref{alg:alg1} summarizes the training process of MCformer.

\section{EXPERIMENTS}
\subsection{Dateset}
We used five datasets containing multivariate time series data to evaluate our model, including Electricity, Traffic, Weather\cite{wu2021autoformer}, Solar-Energy\cite{lai2018lstnet}, PEMS\cite{liu2022scinet}. These datasets are widely used as benchmarks for multivariate time series forecasting, and details about the datasets can be found in Table \ref{dataset}. It is worth noting that datasets like ETT (ETTh1, ETTh2, ETTm1, ETTm2)\cite{zhou2021informer} and ILI have only 7 channels of data. Due to the fewer number of channels, they were not used for model evaluation.

\textbf{Weather}: The weather dataset was collected at approximately 1, 600 locations across the United States between 2010 and 2013, with a sampling frequency of one record every ten minutes. This dataset contains 21 channels.

\textbf{Solar-Energy}: The Solar-Energy dataset documents the solar power generation of a photovoltaic power station in Alabama in 2006, with readings captured every 10 minutes. Data from a total of 137 channels were collected.

\textbf{Electricity}: The Electricity dataset captures the hourly electricity consumption (measured in kilowatt-hours) of 321 customers from 2012 to 2014.

\textbf{Traffic}: The Traffic dataset encompasses road occupancy data recorded by sensors on San Francisco Bay area freeways from 2015 to 2016. Readings are logged on an hourly basis, ranging from 0 to 1. A total of 862 sensor channels are included.

\textbf{PEMS}: The PEMS dataset is collected by the California Department of Transportation's Performance Measurement System (PeMS) in the California region. There are 358 channels of data recorded

\begin{table}[!t]
    \caption{The detailed information of benchmark datasets used for testing}
    \centering
    \tabcolsep=1mm
    \begin{tabular}{cccccc}
    \toprule 
    \multicolumn{1}{c|}{Dataset} & Electricity & Traffic & Weather & Solar-Energy & PEMS \\
    Dataset Detail &  &  &  &  &  \\ \cline{1-1}
    \midrule 
    \multicolumn{1}{c|}{features} & 321 & 862 & 21 & 137 & 358 \\
    \multicolumn{1}{c|}{frequency} & 1h & 1h & 10m & 10m & 5m \\
    \multicolumn{1}{c|}{length} & 26211 & 17451 & 52603 & 52179 & 21352 \\ 
    \bottomrule 
    \end{tabular}
    \label{dataset}
\end{table}
\subsection{Baseline}
In the realm of time series forecasting, deep learning models have made remarkable strides, surpassing traditional approaches in a multitude of tasks. To assess the performance of our proposed methodology, we have meticulously selected a cohort of state-of-the-art (SOTA) multivariate time series forecasting models. Transformer-based models have exhibited exceptional performance in time series forecasting tasks. We have chosen the most representative models from this class, including InFormer\cite{zhou2021informer}, AutoFormer\cite{wu2021autoformer}, FEDFormer\cite{zhou2022fedformer}, CrossFormer\cite{zhang2023crossformer}, and PatchTST\cite{Yuqietal-2023-PatchTST}. Additionally, given the promising results recently achieved by MLP-based models, we have opted to include the most notable representatives, namely DLinear\cite{zeng2023dlinear} and TiDE\cite{das2023tide}. CNN-based models possess distinct advantages in multivariate feature extraction. Consequently, we have incorporated TimesNet\cite{wu2023timesnet} into our evaluation.

\subsection{forecasting and baseline comparison}
\textbf{Setup }We followed the experimental settings of TimesNet, where the look-back window length and forecasting window length for the Solar-Energy, Weather, Traffic, and Electricity datasets were set to 96. The forecasting window length was chosen as 
\(h\in\{96,192,336,720\}\). For the PEMS dataset, the look-back window length was 96, and the forecasting window length was \(h\in\{12,24,48,96\}\).

\begin{table*}[!t]\scriptsize
\caption{Full results on the multivariate forecasting task. We used a look-back window of length 96 for all datasets, and for PEMS, we used forecasting windows \(h\in\{12,24,48,96\}\), while for others \(h\in\{96,192,336,720\}\). The best results are highlighted in \textbf{bold}, and the second-best results are \underline{underlined}.}
\setlength{\tabcolsep}{4.5pt}
\renewcommand{\arraystretch}{1.5}
\centering
    \begin{tabular}{cc|cc|cc|cc|cc|cc|cc|cc|cc|cc}
    \toprule 
    \multicolumn{2}{c|}{\multirow{2}{*}{Models}} & \multicolumn{2}{c|}{MCformer} & \multicolumn{2}{c|}{TiDE\cite{das2023tide}} & \multicolumn{2}{c|}{PatchTST\cite{Yuqietal-2023-PatchTST}} & \multicolumn{2}{c|}{TimesNet\cite{wu2023timesnet}} & \multicolumn{2}{c|}{CrossFormer\cite{zhang2023crossformer}} & \multicolumn{2}{c|}{Dlinear\cite{zeng2023dlinear}} & \multicolumn{2}{c|}{FEDFormer\cite{zhou2022fedformer}} & \multicolumn{2}{c|}{AutoFormer\cite{wu2021autoformer}} & \multicolumn{2}{c}{InFormer\cite{zhou2021informer}} \\
    &  & \multicolumn{2}{c|}{(Ours)} & \multicolumn{2}{c|}{(2023)} & \multicolumn{2}{c|}{(2023)} & \multicolumn{2}{c|}{(2023)} & \multicolumn{2}{c|}{(2023)} & \multicolumn{2}{c|}{(2023)} & \multicolumn{2}{c|}{(2022)} & \multicolumn{2}{c|}{(2021)} & \multicolumn{2}{c}{(2021)} \\ \cline{1-20}
    \multicolumn{2}{c|}{Metric} & MSE & MAE & MSE & MAE & MSE & MAE & MSE & MAE & MSE & MAE & MSE & MAE & MSE & MAE & MSE & MAE & MSE & MAE \\ \cline{1-20}
    \multirow{4}{*}{\rotatebox{90}{Weather}} & 96 & \underline{0.167}  & \textbf{0.213}  & 0.202  & 0.261  & 0.177  & 0.218  & 0.172  & 0.220  & \textbf{0.158}  & \underline{0.230}  & 0.196  & 0.255  & 0.217  & 0.296  & 0.266&0.336&0.300& 0.384  \\
    & 192 & \underline{0.215} & \textbf{0.257}& 0.242  & 0.298  & 0.225  & 0.259  & 0.219  & 0.261  & \textbf{0.206}  & \underline{0.277}  & 0.237  & 0.296  & 0.276  & 0.336  & 0.307  & 0.367  & 0.598  & 0.544  \\
    & 336 & \textbf{0.270}  & \textbf{0.297}  & 0.287  & 0.335  & 0.278  & 0.297  & 0.280  & 0.306  & \underline{0.272}  & \underline{0.335}  & 0.283  & 0.335  & 0.339  & 0.380  & 0.359  & 0.395  & 0.578  & 0.523  \\
    & 720 & \textbf{0.348}  & \textbf{0.348}  & \underline{0.351}  & \underline{0.386}  & 0.354  & 0.348  & 0.365  & 0.359  & 0.398  & 0.418  & 0.345  & 0.381  & 0.403  & 0.428  & 0.419  & 0.428  & 1.059  & 0.741  \\ \cline{1-20}
    \multirow{4}{*}{\rotatebox{90}{Traffic}} & 96 & \textbf{0.433}  & \textbf{0.288} & 0.805  & 0.493  & 0.544  & 0.359  & 0.593  & 0.321  & \underline{0.522}  & \underline{0.290}  & 0.650  & 0.396  & 0.587  & 0.366  & 0.613  & 0.388  & 0.719  & 0.391  \\
    & 192 & \textbf{0.440}  & \textbf{0.286} & 0.756  & 0.474  & 0.540  & 0.354  & 0.617  & 0.336  & \underline{0.530}  & \underline{0.293}  & 0.598  & 0.370  & 0.604  & 0.373  & 0.616  & 0.382  & 0.696  & 0.379  \\
    & 336 & \textbf{0.454}  & \textbf{0.292}  & 0.762  & 0.477  & \underline{0.551}  & \underline{0.358}  & 0.629  & 0.336  & 0.558  & 0.305  & 0.605  & 0.373  & 0.621  & 0.383  & 0.622  & 0.337  & 0.777  & 0.420  \\
    & 720 & \textbf{0.489} &\textbf{0.312}  & 0.719&0.449&\underline{0.586}&\underline{ 0.375}&0.640&0.350& 0.589&0.328& 0.645  & 0.394  & 0.626  & 0.382  & 0.660  & 0.408  & 0.864  & 0.472  \\ \cline{1-20}
    \multirow{4}{*}{\rotatebox{90}{Electricity}} & 96 & \textbf{0.163}  & \textbf{0.255}  & 0.237  & 0.329  & 0.195  & 0.285  & \underline{0.168}  & \underline{0.272}  & 0.219  & 0.314  & 0.197  & 0.282  & 0.193  & 0.308  & 0.201  & 0.317  & 0.274  & 0.368  \\
    & 192 & \textbf{0.172}  & \textbf{0.262}  & 0.236  & 0.330  & 0.199  & 0.289  & \underline{0.184}  & \underline{0.289}  & 0.231  & 0.322  & 0.196  & 0.285  & 0.201  & 0.315  & 0.222  & 0.334  & 0.296  & 0.386  \\
    & 336 & \textbf{0.190}  & \textbf{0.279}  & 0.249  & 0.344  & 0.215  & 0.305  & \underline{0.198}  & \underline{0.300}  & 0.246  & 0.337  & 0.209  & 0.301  & 0.214  & 0.329  & 0.231  & 0.338  & 0.300  & 0.394  \\
     & 720 & \underline{0.229}  & \underline{0.311}  & 0.284  & 0.373  & 0.256  & 0.337  & \textbf{0.220}  & \textbf{0.320}  & 0.280  & 0.363  & 0.245  & 0.333  & 0.246  & 0.355  & 0.254  & 0.361  & 0.373  & 0.439  \\ \cline{1-20}
    \multirow{4}{*}{\rotatebox{90}{Solar-Energy}} & 96 & \textbf{0.212}  & \textbf{0.256}  & 0.312  & 0.399  & \underline{0.234}  & 0.286  & 0.250  & 0.292  & 0.310  & 0.331  & 0.290  & 0.378  & 0.242  & 0.342  & 0.884  & 0.711  & 0.236  & \underline{0.259}  \\
     & 192 & \underline{0.241}  & \underline{0.275}  & 0.339  & 0.416  & 0.267  & 0.310  & 0.296  & 0.318  & 0.734  & 0.725  & 0.320  & 0.398  & 0.285  & 0.380  & 0.834  & 0.692  & \textbf{0.217}  & \textbf{0.269 } \\
     & 336 & \underline{0.258}  & \underline{0.287}  & 0.368  & 0.430  & 0.290  & 0.315  & 0.319  & 0.330  & 0.750  & 0.735  & 0.353  & 0.415  & 0.282  & 0.376  & 0.941  & 0.723  & \textbf{0.249}  & \textbf{0.283}  \\
     & 720 & \underline{0.264}  & \textbf{0.296}  & 0.370  & 0.425  & 0.289  & 0.317  & 0.338  & 0.337  & 0.769  & 0.765  & 0.356  & 0.413  & 0.357  & 0.427  & 0.882  & 0.717  & \textbf{0.241}  & \underline{0.317}  \\ \cline{1-20}
    \multirow{4}{*}{\rotatebox{90}{PEMS}} & 12 & \textbf{0.072}  & \textbf{0.180}  & 0.178  & 0.305  & 0.099  & 0.216  & \underline{0.085}  & \underline{0.192}  & 0.090  & 0.203  & 0.122  & 0.243  & 0.126  & 0.251  & 0.272  & 0.385  & 0.126  & 0.233  \\
    & 24 & \textbf{0.100}  & \textbf{0.212}  & 0.257  & 0.371  & 0.142  & 0.259  & \underline{0.118}  & \underline{0.223}  & 0.121  & 0.240  & 0.201  & 0.317  & 0.149  & 0.275  & 0.334  & 0.440  & 0.139  & 0.250  \\
    & 48 & \underline{0.164}  & \underline{0.271}  & 0.379  & 0.463  & 0.211  & 0.319  & \textbf{0.155}  & \textbf{0.260}  & 0.202  & 0.317  & 0.333  & 0.425  & 0.227  & 0.348  & 1.032  & 0.782  & 0.186  & 0.289  \\
    & 96 & \underline{0.240}  & \underline{0.337}  & 0.490  & 0.539  & 0.269  & 0.370  & \textbf{0.228}  & \textbf{0.317 } & 0.262  & 0.367  & 0.457  & 0.515  & 0.348  & 0.434  & 1.031  & 0.796  & 0.233  & 0.323  \\
    \bottomrule 
    \end{tabular}
    \label{main result}
\end{table*}

\textbf{Results} As shown in the Table\ref{main result}, our proposed MCFormer model achieved the best performance across all datasets. In all experimental comparisons, we secured 12 first places and 8 second places in MSE, and 15 first places and 5 second places in MAE. This indicates that our method outperforms all compared methods. Notably, compared to the single-channel strategies PatchTST and TiDE, our method shows significant improvements. This suggests that our approach effectively captures inter-channel dependencies, leading to additional performance gains. In comparison to the all-channel strategies TimesNet and Crossformer, our MCFormer effectively mitigates the disruption of multi-channel dependencies on long-term information.
\begin{figure}[!htbp]
    \centering
    \includegraphics[width=1\linewidth]{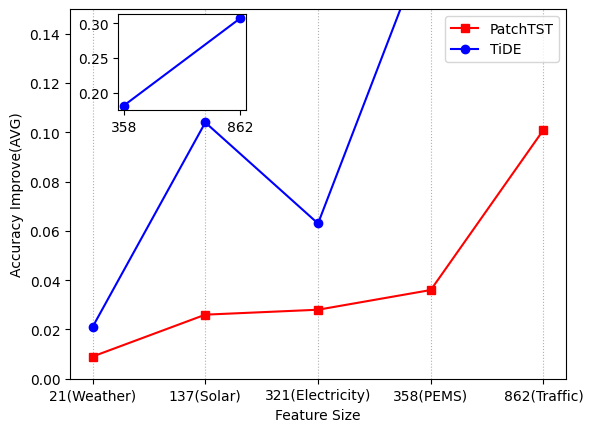}
    \caption{Accuracy improvement versus number of features. We computed the average MSE improvement of MCformer compared to the single-channel strategies TiDE and PatchTST across different channel numbers. The results indicate that as the number of channels increases, the performance improvement of MCformer gradually becomes more significant. This suggests that MCformer can effectively capture dependencies between Multivariate data, thereby enhancing predictive performance.}
    \label{avg_imporve}
\end{figure}

\subsection{Ablation study}
In this section, we designed two ablation experiments to investigate the impact of different channel fusion quantities on the model's performance and the effectiveness of the channel fusion method on datasets with varying channel numbers.

\begin{figure*}[!t]
\centering
    \subfloat
    {
        \begin{minipage}[b]{.3\linewidth}
            \centering
            \includegraphics[scale=0.28]{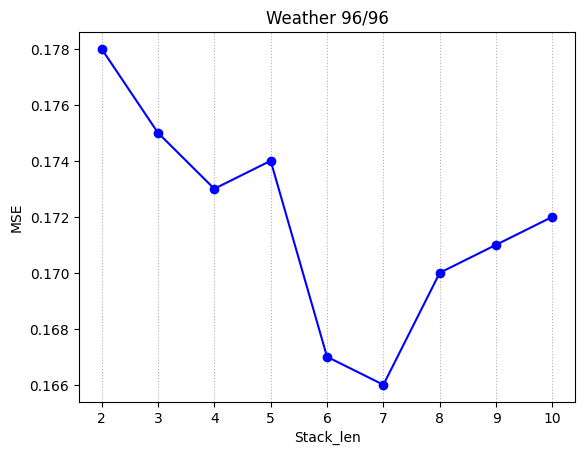}
        \end{minipage}
    }
    \subfloat
    {
     	\begin{minipage}[b]{.3\linewidth}
            \centering
            \includegraphics[scale=0.28]{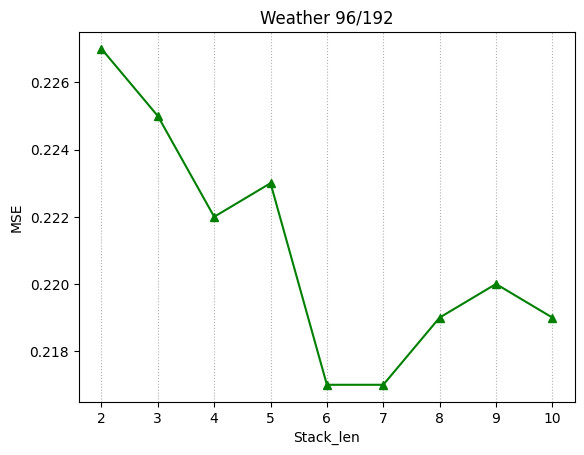}
        \end{minipage}
    }
    \subfloat
    {
     	\begin{minipage}[b]{.3\linewidth}
            \centering
            \includegraphics[scale=0.28]{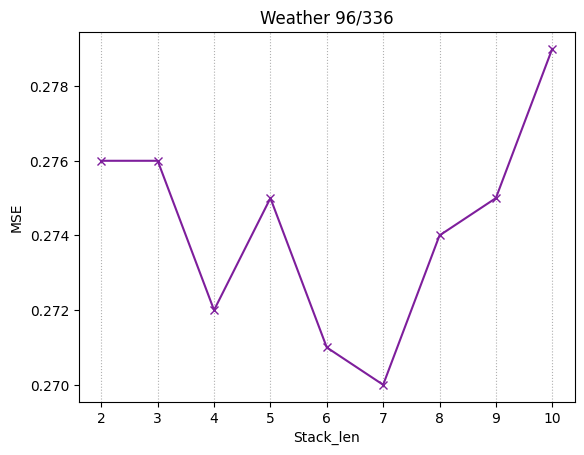}
        \end{minipage}
    }
    \\
    \subfloat
    {
        \begin{minipage}[b]{.3\linewidth}
            \centering
            \includegraphics[scale=0.28]{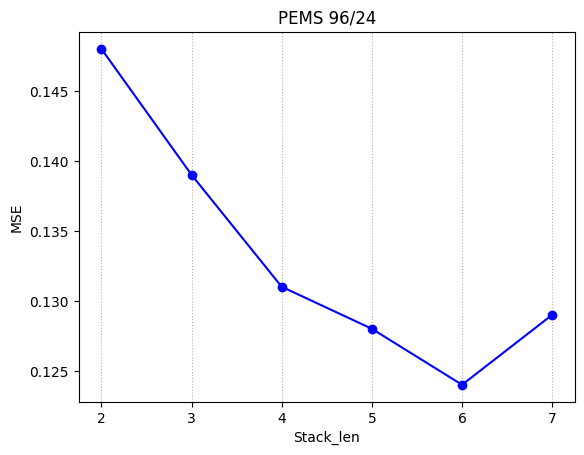}
        \end{minipage}
    }
    \subfloat
    {
     	\begin{minipage}[b]{.3\linewidth}
            \centering
            \includegraphics[scale=0.28]{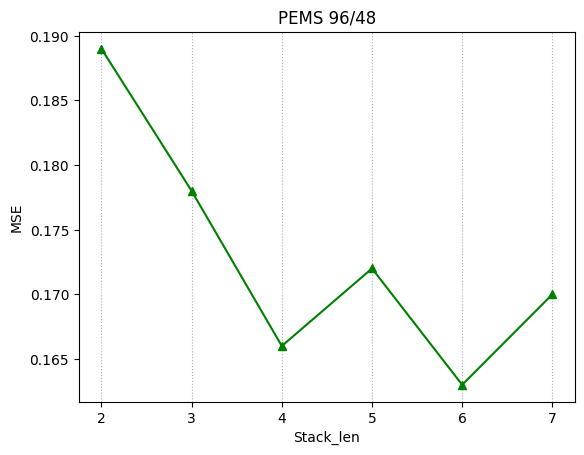}
        \end{minipage}
    }
    \subfloat
    {
     	\begin{minipage}[b]{.3\linewidth}
            \centering
            \includegraphics[scale=0.28]{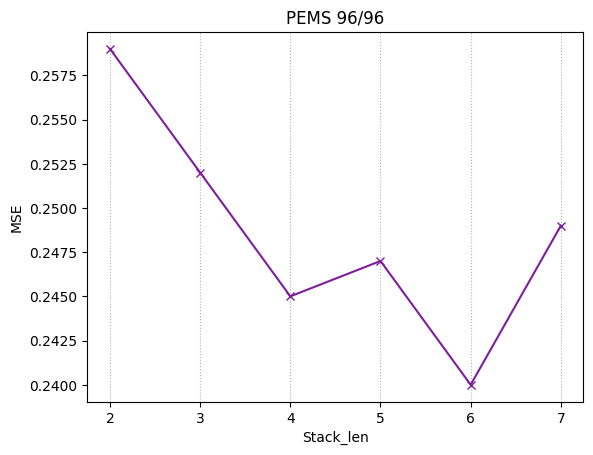}
        \end{minipage}
    }\\
    \subfloat
    {
        \begin{minipage}[b]{.3\linewidth}
            \centering
            \includegraphics[scale=0.28]{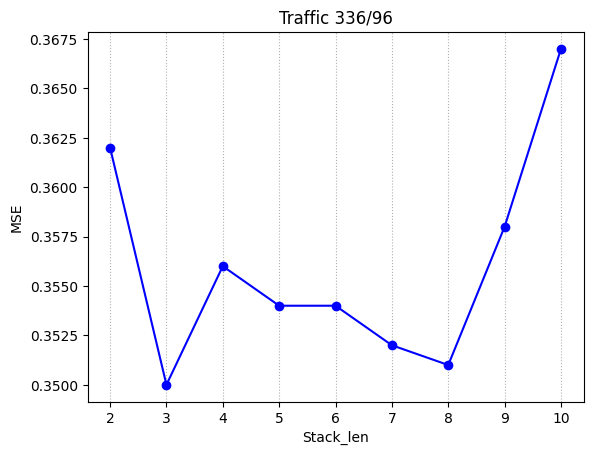}
        \end{minipage}
    }
    \subfloat
    {
     	\begin{minipage}[b]{.3\linewidth}
            \centering
            \includegraphics[scale=0.28]{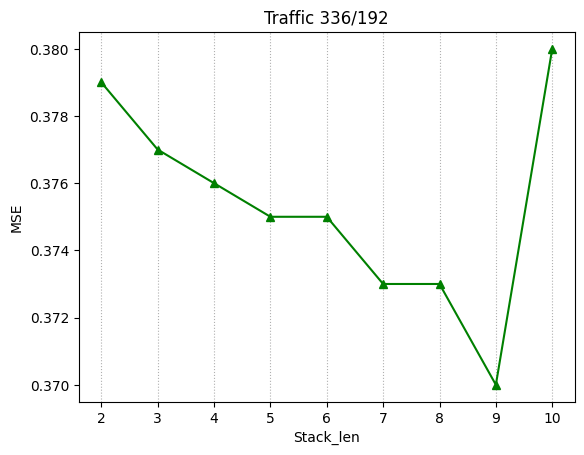}
        \end{minipage}
    }
    \subfloat
    {
     	\begin{minipage}[b]{.3\linewidth}
            \centering
            \includegraphics[scale=0.28]{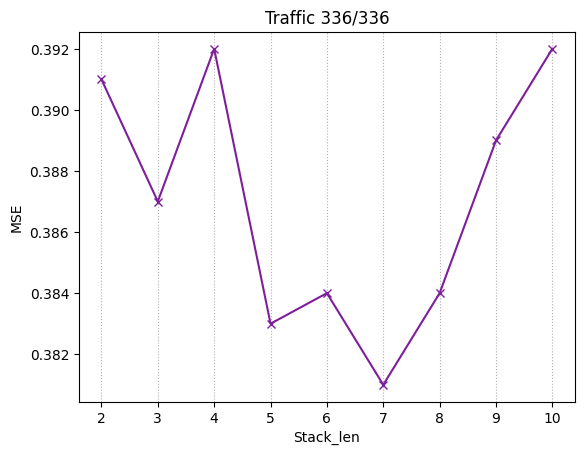}
        \end{minipage}
    }
\caption{Forecasting performance (MSE) of Weather, PEMS and Traffic datasets with different look-back window and forecasting window.For Traffic, the look-back window is 336, while for others, it's 96. We use forecasting lengths \(h\in\{24,48,96\}\) for PEMS,and \(h\in\{96,192,336\}\) for others.  }
\label{channel fusion quantity}
\end{figure*}

\textbf{Impact on Data with Different Channel Numbers} We hypothesize that under a single-channel strategy, the model may experience the issue of forgetting inter-channel correlations when the number of channels is high. To validate this hypothesis, we designed experiments to test the predictive performance of the model under different channel numbers. We selected the Electricity, Traffic, Weather, Solar-Energy, and PEMS datasets for experimentation, with their characteristics detailed in Table \ref{dataset}. We conducted experiments on different datasets with varying forecasting lengths and calculated their average performances. In order to compare the performance improvement of our model using a single-channel strategy, we contrasted TiDE and PatchTST. As depicted in Fig. \ref{avg_imporve}, it is evident that MCformer exhibits a more significant enhancement in average MSE as the number of channels increases when compared to the single-channel strategy model. This indicates that our model is capable of effectively addressing the issue of inter-channel correlation forgetting and does not compromise model performance with an increase in channels.

\textbf{Channel Fusion Quantity} Blending channels may compromise the long-term features of the sequence, as demonstrated by PatchTST and PETFormer. Therefore, we need to investigate the impact of channel quantity on model performance. In Fig. \ref{channel fusion quantity}, it is evident that model performance improves with a certain number of channels. However, on the Traffic, when the number of channels reaches 9, model performance starts to decline. When the number of channels reaches 10, the performance of blended channels even falls below that of the single-channel strategy. Similar trends are observed in the PEMS and Weather datasets, where the MSE initially decreases and then increases as the amount of mixed channel data grows. This suggests that our model effectively learns inter-channel correlations while minimizing the disruption of the model's ability to capture long-term features.

\subsection{Visualiztion Analysis}
\begin{figure*}
    \centering
    \includegraphics[width=1\linewidth]{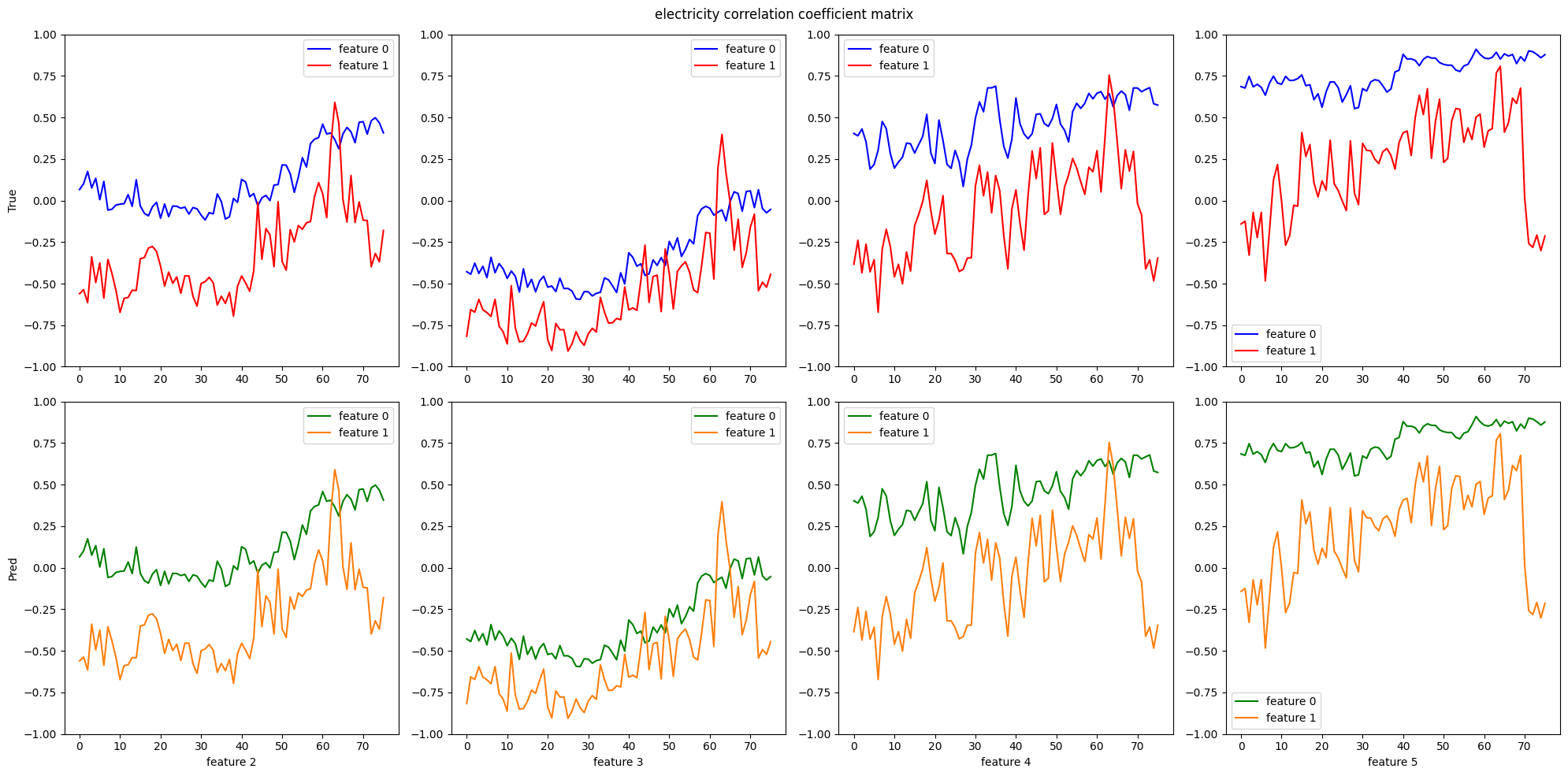}
    \caption{Visualization of correlation variation of certain characteristics of Electricity dataset. Selected a subset of features from the Electricity dataset for temporal correlation analysis. We randomly sampled 6 features from the Electricity dataset. The four subplots above show the correlation changes between features 0 and 1 and features 2, 3, 4, and 5 in the real values. The predicted values are shown below. As can be clearly seen in the figure, the temporal correlation of different features is non-stationary. Our model, however, fits the correlation changes of the real values very well in terms of prediction values.}
    \label{correlation}
\end{figure*}

Correlation analysis is a crucial method for measuring the degree of relationship between two variables, widely applied in the field of time series analysis. We attempt to analyze and evaluate the correlation of channel-wise information in the dataset by visualizing the correlation changes, which provides an explanation for the effectiveness of our model. The temporal variation of inter-channel correlation reflects changes in the correlated features among multiple channels. As shown in the Fig. \ref{correlation}, we utilize the same look-back window size as the forecasting to analyze the inter-channel correlation in the time series. Since the number of channels in the dataset exceeds 20, we randomly sampled channels for analysis. In the Electricity dataset, we observe that the inter-channel correlation changes over time, and there are significant variations in the correlation among some channels. This indicates that, over time, inter-channel dependency information, like long-term information, exhibits diversity and requires increased attention during modeling. The comparison between the real and predicted values shows that our model fits the correlation changes between the real values very well.

\section{FUTURE WORK}

Based on the current research, our future work can evolve in two directions to further advance the field of multi-channel time-series forecasting. Firstly, we can explore more complex Mixed Channels strategies by finely tuning the combination of channels to further enhance the model's performance in handling multi-channel time-series data. This may involve in-depth analysis of channel correlations and optimization of model architecture to achieve more accurate long-term feature modeling.

Furthermore, interpretability and explainability of the model are also crucial directions for future research. Enhancing our understanding of the model's focus on different channels and its decision-making process during forecasting can increase trust in the model's forecast results, improving its acceptability in practical applications.

In summary, future work should delve deeper into refining Mixed Channels strategies and enhancing model interpretability. This will contribute to further advancing and applying multi-channel time-series forecasting in various domains.

\section{CONCLUSION}

In this paper, we propose a Multivariate time series forecasting model, MCformer, which leverages a mixed Multivariate feature. MCformer effectively addresses the issue of performance degradation caused by the disruption of channel information to long-term features by incorporating a limited set of mixed multi-channel data while preserving the advantages of the single-channel strategy in expanding the dataset. In experiments, MCformer consistently outperformed other models across all datasets. Furthermore, we conducted an in-depth analysis of the impact of the number of fused channels on the model. In the future, we plan to further explore the various effects of multi-channel features on time series analysis tasks.

\bibliographystyle{IEEEtran}
\bibliography{reference}

\begin{thebibliography}{10}
\providecommand{\url}[1]{#1}
\csname url@samestyle\endcsname
\providecommand{\newblock}{\relax}
\providecommand{\bibinfo}[2]{#2}
\providecommand{\BIBentrySTDinterwordspacing}{\spaceskip=0pt\relax}
\providecommand{\BIBentryALTinterwordstretchfactor}{4}
\providecommand{\BIBentryALTinterwordspacing}{\spaceskip=\fontdimen2\font plus
\BIBentryALTinterwordstretchfactor\fontdimen3\font minus
  \fontdimen4\font\relax}
\providecommand{\BIBforeignlanguage}[2]{{%
\expandafter\ifx\csname l@#1\endcsname\relax
\typeout{** WARNING: IEEEtran.bst: No hyphenation pattern has been}%
\typeout{** loaded for the language `#1'. Using the pattern for}%
\typeout{** the default language instead.}%
\else
\language=\csname l@#1\endcsname
\fi
#2}}
\providecommand{\BIBdecl}{\relax}
\BIBdecl

\bibitem{angryk2020multivariate}
R.~A. Angryk, P.~C. Martens, B.~Aydin, D.~Kempton, S.~S. Mahajan, S.~Basodi,
  A.~Ahmadzadeh, X.~Cai, S.~Filali~Boubrahimi, S.~M. Hamdi \emph{et~al.},
  ``Multivariate time series dataset for space weather data analytics,''
  \emph{Scientific data}, vol.~7, no.~1, p. 227, 2020.

\bibitem{han2021joint}
J.~Han, H.~Liu, H.~Zhu, H.~Xiong, and D.~Dou, ``Joint air quality and weather
  prediction based on multi-adversarial spatiotemporal networks,'' in
  \emph{Proceedings of the AAAI Conference on Artificial Intelligence},
  vol.~35, no.~5, 2021, pp. 4081--4089.

\bibitem{cheng2018deepiot}
Y.~Cheng, S.~Wan, K.-K.~R. Choo \emph{et~al.}, ``Deep belief network for
  meteorological time series prediction in the internet of things,'' \emph{IEEE
  internet of things journal}, vol.~6, no.~3, pp. 4369--4376, 2018.

\bibitem{ghosh2009multivariate}
B.~Ghosh, B.~Basu, and M.~O'Mahony, ``Multivariate short-term traffic flow
  forecasting using time-series analysis,'' \emph{IEEE transactions on
  intelligent transportation systems}, vol.~10, no.~2, pp. 246--254, 2009.

\bibitem{cirstea2022traffic2}
R.-G. Cirstea, B.~Yang, C.~Guo, T.~Kieu, and S.~Pan, ``Towards spatio-temporal
  aware traffic time series forecasting,'' in \emph{2022 IEEE 38th
  International Conference on Data Engineering (ICDE)}.\hskip 1em plus 0.5em
  minus 0.4em\relax IEEE, 2022, pp. 2900--2913.

\bibitem{qin2023traffic3}
Y.~Qin, H.~Luo, F.~Zhao, Y.~Fang, X.~Tao, and C.~Wang, ``Spatio-temporal
  hierarchical mlp network for traffic forecasting,'' \emph{Information
  Sciences}, vol. 632, pp. 543--554, 2023.

\bibitem{zhou2020reinforced}
F.~Zhou, Q.~Yang, K.~Zhang, G.~Trajcevski, T.~Zhong, and A.~Khokhar,
  ``Reinforced spatiotemporal attentive graph neural networks for traffic
  forecasting,'' \emph{IEEE Internet of Things Journal}, vol.~7, no.~7, pp.
  6414--6428, 2020.

\bibitem{stefenon2023wavelet}
S.~F. Stefenon, L.~O. Seman, L.~S. Aquino, and L.~dos Santos~Coelho,
  ``Wavelet-seq2seq-lstm with attention for time series forecasting of level of
  dams in hydroelectric power plants,'' \emph{Energy}, vol. 274, p. 127350,
  2023.

\bibitem{han2021multivariateiot}
J.~Han, G.~H. Lee, S.~Park, J.~Lee, and J.~K. Choi, ``A
  multivariate-time-series-prediction-based adaptive data transmission period
  control algorithm for iot networks,'' \emph{IEEE Internet of Things Journal},
  vol.~9, no.~1, pp. 419--436, 2021.

\bibitem{hua2023kae}
Q.~Hua, D.~Yang, S.~Qian, H.~Hu, J.~Cao, and G.~Xue, ``Kae-informer: A
  knowledge auto-embedding informer for forecasting long-term workloads of
  microservices,'' in \emph{Proceedings of the ACM Web Conference 2023}, 2023,
  pp. 1551--1561.

\bibitem{yi2023frequency}
K.~Yi, Q.~Zhang, W.~Fan, S.~Wang, P.~Wang, H.~He, N.~An, D.~Lian, L.~Cao, and
  Z.~Niu, ``Frequency-domain {MLP}s are more effective learners in time series
  forecasting,'' in \emph{Thirty-seventh Conference on Neural Information
  Processing Systems}, 2023.

\bibitem{ni2023basisformer}
Z.~Ni, H.~Yu, S.~Liu, J.~Li, and W.~Lin, ``{Basisformer}: Attention-based time
  series forecasting with learnable and interpretable basis,'' in
  \emph{Advances in Neural Information Processing Systems}, 2023.

\bibitem{li2019logtrans}
S.~Li, X.~Jin, Y.~Xuan, X.~Zhou, W.~Chen, Y.-X. Wang, and X.~Yan, ``Enhancing
  the locality and breaking the memory bottleneck of transformer on time series
  forecasting,'' \emph{Advances in neural information processing systems},
  vol.~32, 2019.

\bibitem{zhou2021informer}
H.~Zhou, S.~Zhang, J.~Peng, S.~Zhang, J.~Li, H.~Xiong, and W.~Zhang,
  ``Informer: Beyond efficient transformer for long sequence time-series
  forecasting,'' in \emph{Proceedings of the AAAI conference on artificial
  intelligence}, vol.~35, no.~12, 2021, pp. 11\,106--11\,115.

\bibitem{kitaev2020reformer}
N.~Kitaev, L.~Kaiser, and A.~Levskaya, ``Reformer: The efficient transformer,''
  in \emph{International Conference on Learning Representations}, 2019.

\bibitem{wu2021autoformer}
H.~Wu, J.~Xu, J.~Wang, and M.~Long, ``Autoformer: Decomposition transformers
  with auto-correlation for long-term series forecasting,'' \emph{Advances in
  Neural Information Processing Systems}, vol.~34, pp. 22\,419--22\,430, 2021.

\bibitem{zhou2022fedformer}
T.~Zhou, Z.~Ma, Q.~Wen, X.~Wang, L.~Sun, and R.~Jin, ``Fedformer: Frequency
  enhanced decomposed transformer for long-term series forecasting,'' in
  \emph{International Conference on Machine Learning}.\hskip 1em plus 0.5em
  minus 0.4em\relax PMLR, 2022, pp. 27\,268--27\,286.

\bibitem{shabani2022scaleformer}
M.~A. Shabani, A.~H. Abdi, L.~Meng, and T.~Sylvain, ``Scaleformer: Iterative
  multi-scale refining transformers for time series forecasting,'' in \emph{The
  Eleventh International Conference on Learning Representations}, 2022.

\bibitem{liu2022pyraformer}
S.~Liu, H.~Yu, C.~Liao, J.~Li, W.~Lin, A.~X. Liu, and S.~Dustdar, ``Pyraformer:
  Low-complexity pyramidal attention for long-range time series modeling and
  forecasting,'' in \emph{International Conference on Learning
  Representations}, 2022.

\bibitem{chen2021fpp}
Z.~Chen, D.~Chen, X.~Zhang, Z.~Yuan, and X.~Cheng, ``Learning graph structures
  with transformer for multivariate time-series anomaly detection in iot,''
  \emph{IEEE Internet of Things Journal}, vol.~9, no.~12, pp. 9179--9189, 2021.

\bibitem{zhang2023crossformer}
Y.~Zhang and J.~Yan, ``Crossformer: Transformer utilizing cross-dimension
  dependency for multivariate time series forecasting,'' in \emph{International
  Conference on Learning Representations}, 2023.

\bibitem{zhang2023sageformer}
Z.~Zhang, X.~Wang, and Y.~Gu, ``Sageformer: Series-aware graph-enhanced
  transformers for multivariate time series forecasting,'' \emph{arXiv preprint
  arXiv:2307.01616}, 2023.

\bibitem{zeng2023dlinear}
A.~Zeng, M.~Chen, L.~Zhang, and Q.~Xu, ``Are transformers effective for time
  series forecasting?'' in \emph{Proceedings of the AAAI conference on
  artificial intelligence}, vol.~37, no.~9, 2023, pp. 11\,121--11\,128.

\bibitem{Yuqietal-2023-PatchTST}
Y.~Nie, N.~H.~Nguyen, P.~Sinthong, and J.~Kalagnanam, ``A time series is worth
  64 words: Long-term forecasting with transformers,'' in \emph{International
  Conference on Learning Representations}, 2023.

\bibitem{han2023capacity}
L.~Han, H.~Ye, and D.~Zhan, ``The capacity and robustness trade-off: Revisiting
  the channel independent strategy for multivariate time series forecasting,''
  \emph{CoRR}, vol. abs/2304.05206, 2023.

\bibitem{lin2023petformer}
S.~Lin, W.~Lin, W.~Wu, S.~Wang, and Y.~Wang, ``Petformer: Long-term time series
  forecasting via placeholder-enhanced transformer,'' \emph{arXiv preprint
  arXiv:2308.04791}, 2023.

\bibitem{wang2023csformer}
H.~Wang, Y.~Mo, N.~Yin, H.~Dai, B.~Li, S.~Fan, and S.~Mo, ``Dance of channel
  and sequence: An efficient attention-based approach for multivariate time
  series forecasting,'' \emph{arXiv preprint arXiv:2312.06220}, 2023.

\bibitem{liu2023itransformer}
Y.~Liu, T.~Hu, H.~Zhang, H.~Wu, S.~Wang, L.~Ma, and M.~Long, ``itransformer:
  Inverted transformers are effective for time series forecasting,''
  \emph{arXiv preprint arXiv:2310.06625}, 2023.

\bibitem{das2023tide}
A.~Das, W.~Kong, A.~Leach, R.~Sen, and R.~Yu, ``Long-term forecasting with
  tide: Time-series dense encoder,'' \emph{arXiv preprint arXiv:2304.08424},
  2023.

\bibitem{lipton2015critical}
Z.~C. Lipton, J.~Berkowitz, and C.~Elkan, ``A critical review of recurrent
  neural networks for sequence learning,'' \emph{arXiv preprint
  arXiv:1506.00019}, 2015.

\bibitem{bartholomew1971arima}
D.~J. Bartholomew, ``Time series analysis forecasting and control.'' 1971.

\bibitem{lai2018lstnet}
G.~Lai, W.-C. Chang, Y.~Yang, and H.~Liu, ``Modeling long-and short-term
  temporal patterns with deep neural networks,'' in \emph{The 41st
  international ACM SIGIR conference on research \& development in information
  retrieval}, 2018, pp. 95--104.

\bibitem{salinas2020deepar}
D.~Salinas, V.~Flunkert, J.~Gasthaus, and T.~Januschowski, ``Deepar:
  Probabilistic forecasting with autoregressive recurrent networks,''
  \emph{International Journal of Forecasting}, vol.~36, no.~3, pp. 1181--1191,
  2020.

\bibitem{wang2022micn}
H.~Wang, J.~Peng, F.~Huang, J.~Wang, J.~Chen, and Y.~Xiao, ``Micn: Multi-scale
  local and global context modeling for long-term series forecasting,'' in
  \emph{The Eleventh International Conference on Learning Representations},
  2022.

\bibitem{wu2023timesnet}
H.~Wu, T.~Hu, Y.~Liu, H.~Zhou, J.~Wang, and M.~Long, ``Timesnet: Temporal
  2d-variation modeling for general time series analysis,'' in
  \emph{International Conference on Learning Representations}, 2023.

\bibitem{dosovitskiy2020vit}
A.~Dosovitskiy, L.~Beyer, A.~Kolesnikov, D.~Weissenborn, X.~Zhai,
  T.~Unterthiner, M.~Dehghani, M.~Minderer, G.~Heigold, S.~Gelly \emph{et~al.},
  ``An image is worth 16x16 words: Transformers for image recognition at
  scale,'' in \emph{International Conference on Learning Representations},
  2020.

\bibitem{gao2023client}
J.~Gao, W.~Hu, and Y.~Chen, ``Client: Cross-variable linear integrated enhanced
  transformer for multivariate long-term time series forecasting,'' \emph{arXiv
  preprint arXiv:2305.18838}, 2023.

\bibitem{kim2021revin}
\BIBentryALTinterwordspacing
T.~Kim, J.~Kim, Y.~Tae, C.~Park, J.-H. Choi, and J.~Choo, ``Reversible instance
  normalization for accurate time-series forecasting against distribution
  shift,'' in \emph{International Conference on Learning Representations},
  2021. [Online]. Available: \url{https://openreview.net/forum?id=cGDAkQo1C0p}
\BIBentrySTDinterwordspacing

\bibitem{vaswani2017attention}
A.~Vaswani, N.~Shazeer, N.~Parmar, J.~Uszkoreit, L.~Jones, A.~N. Gomez,
  {\L}.~Kaiser, and I.~Polosukhin, ``Attention is all you need,''
  \emph{Advances in neural information processing systems}, vol.~30, 2017.

\bibitem{liu2022scinet}
M.~Liu, A.~Zeng, M.~Chen, Z.~Xu, Q.~Lai, L.~Ma, and Q.~Xu, ``Scinet: Time
  series modeling and forecasting with sample convolution and interaction,''
  \emph{Advances in Neural Information Processing Systems}, vol.~35, pp.
  5816--5828, 2022.

\end{thebibliography}

\vspace{11pt}

\begin{IEEEbiography}[{\includegraphics[width=1in,height=1.25in,clip,keepaspectratio]{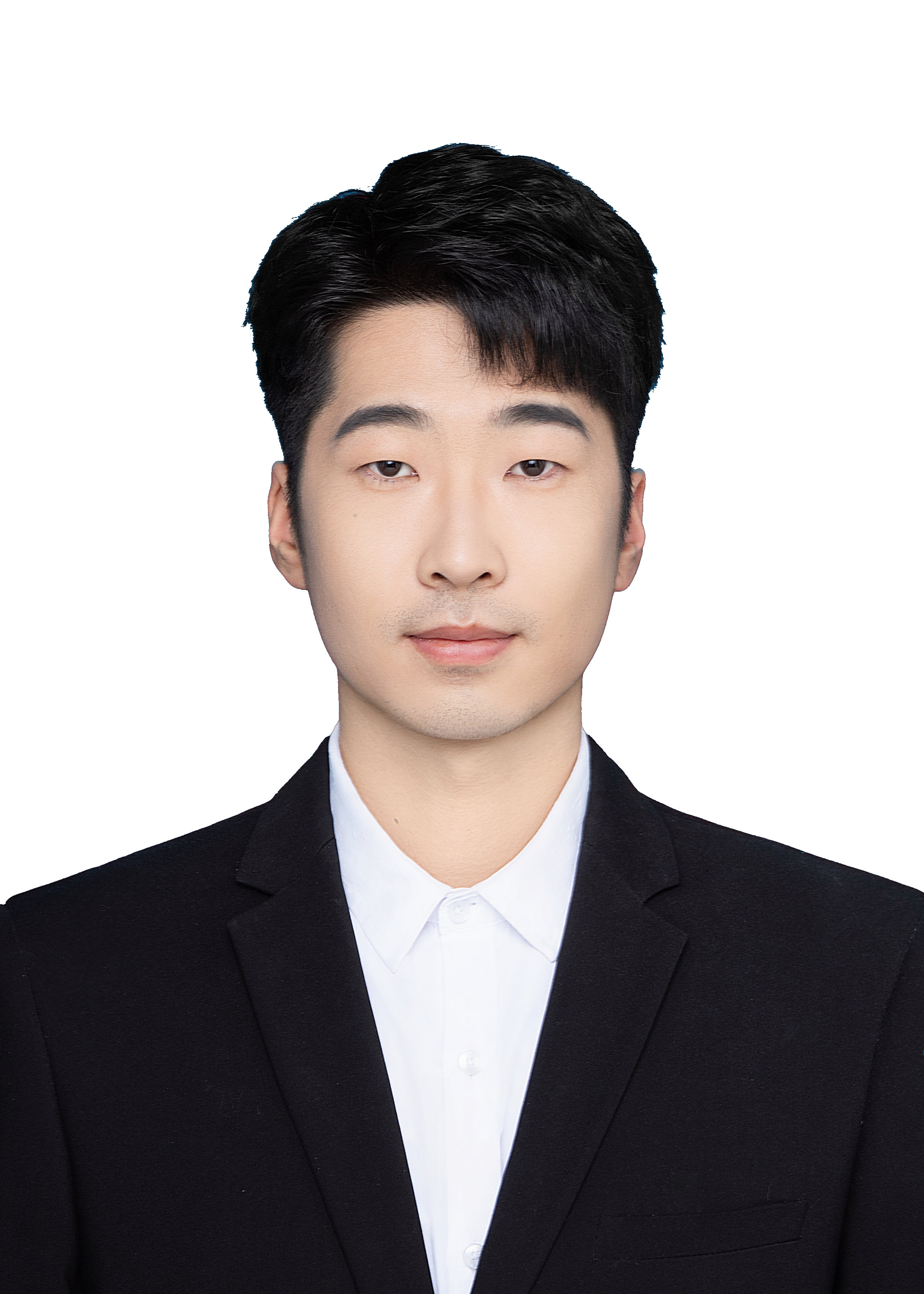}}]{Wenyong Han}received B.S. degree from the School of Computer Science, University of South Chinain 2020. He has been pursuing his Master's degree in the School of Computer Science,  University of South China since 2022.\\
His current research interests include neural networks and time series analysis.
\end{IEEEbiography}

\vspace{11pt}

\begin{IEEEbiography}[{\includegraphics[width=1in,height=1.25in,
  clip,keepaspectratio]{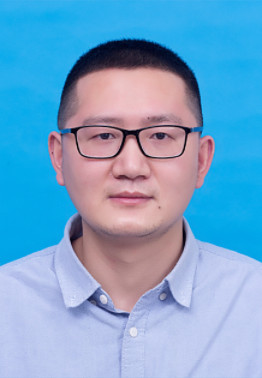}}]{Tao Zhu}
 received his B.E. degree from Central South University, Changsha, China, and Ph.D. from University of Science and Technology of China, Hefei, China, in 2009 and 2015 respectively. He is currently an associate professor at University of South China, Hengyang, China. He is the principal investigator of several projects funded by the National Natural Science Foundation of China and Science Foundation of Hunan Province etc. He is now the Chair of IEEE CIS Smart World Technical Committee Task Force on "User-Centred Smart Systems". His research interests include IoT, pervasive computing, assisted living and evolutionary computation.
\end{IEEEbiography}

\begin{IEEEbiography}[{\includegraphics[width=1in,height=1.25in,
  clip,keepaspectratio]{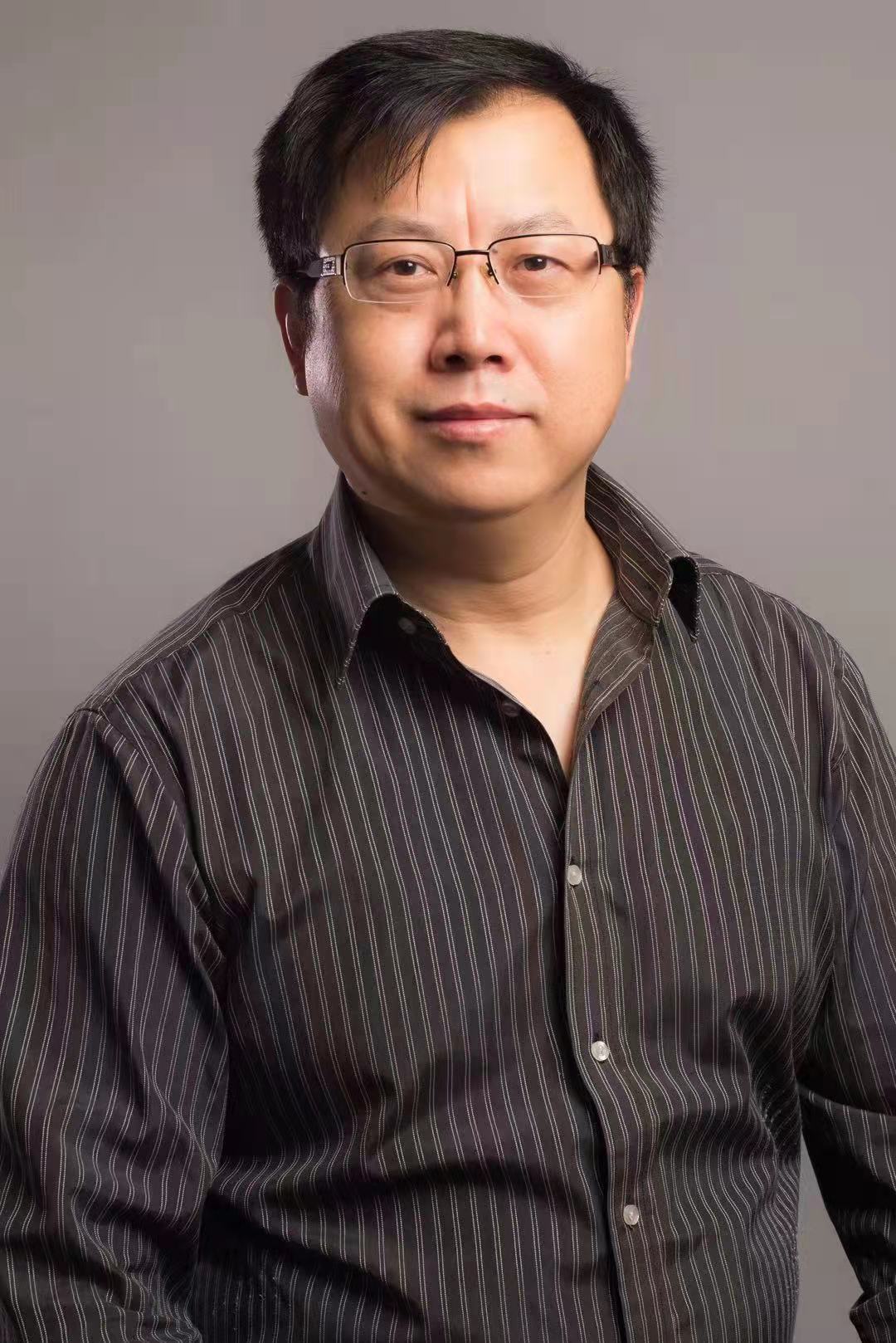}}]{Liming  Chen} is a Professor of School of Computer Science and Technology, Dalian University of Technology, China. His current research interests include pervasive computing, data analytics, artificial intelligence and user centered intelligent systems and their applications in health care and cybersecurity. He has published over 250 papers in the aforementioned areas.
 Liming is an IET Fellow and a Senior Member of IEEE.
\end{IEEEbiography}

\begin{IEEEbiography}[{\includegraphics[width=1in,height=1.25in,
  clip,keepaspectratio]{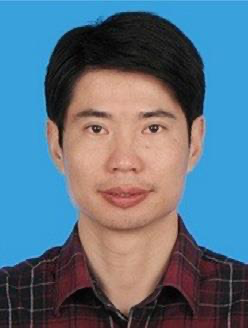}}]{Huansheng Ning}
 received his B.S. degree from
 Anhui University in 1996 and his Ph.D. degree
 from Beihang University in 2001. He is currently
 a Professor and Vice Dean with the School of
 Computer and Communication Engineering, University of Science and Technology Beijing and China
 and Beijing Engineering Research Center for Cyberspace Data Analysis and Applications, China,
 and the founder and principal at the Cybermatics and
 Cyberspace International Science and Technology
 Cooperation Base. He has authored several books
 and over 70 papers in journals and at international conferences/workshops.
 He has been the Associate Editor of the IEEE Systems Journal and IEEE  Internet
of Things Journal, Chairman (2012) and Executive Chairman (2013) of the
 program committee at the IEEE International Internet of Things Conference,
 and the Co-Executive Chairman of the 2013 International Cyber Technology
 Conference and the 2015 Smart World Congress. His awards include the
 IEEE Computer Society Meritorious Service Award and the IEEE Computer
 Society Golden Core Member Award. His current research interests include
the  Internet of Things, Cyber Physical Social Systems, electromagnetic sensing
 and computing.
 
\end{IEEEbiography}

\begin{IEEEbiography}[{\includegraphics[width=1in,height=1.25in,
  clip,keepaspectratio]{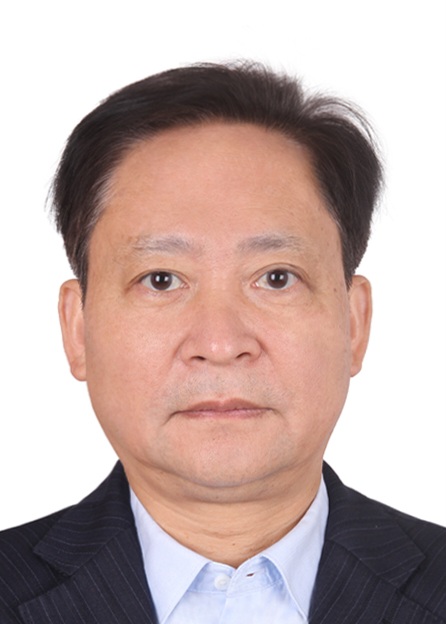}}]{Yang Luo}
 attained a Bachelor's degree in Mathematics from the Department of Mathematics at Hunan Normal University in 1983. Subsequently, in 2005, he earned a Master's degree in Software Engineering from the School of Information Science and Engineering at Central South University,Hunan, China. Presently, he holds the position of Professor at University of South China, Hengyang, China , specializing primarily in the fields of image processing and software engineering.
\end{IEEEbiography}

\begin{IEEEbiography}[{\includegraphics[width=1in,height=1.25in,
  clip,keepaspectratio]{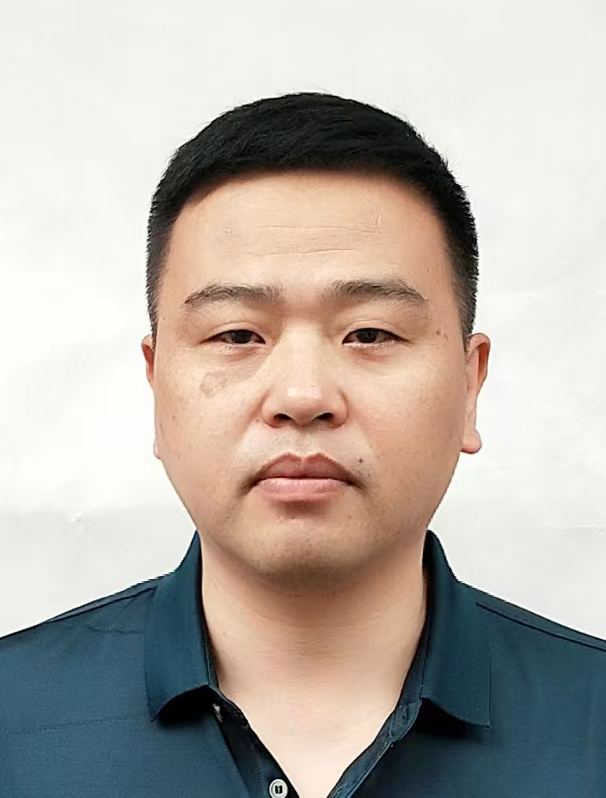}}]{Yaping Wan}
 received his B.S. degree from Huazhong University of Science\&Techno logy(HUST) in 2004 and his Ph.D. degree from HUST in 2009. He is currently a Professor and Dean with the School of Computer, University of South China and the International Cooperation Research Center for Medical Big Data of Hunan Province. He has authored several books and over 40 papers in journals and at international conferences/workshops. He has been the Workshop Chairman (2022) at the 16th IEEE International Conference on Big Data Science and Engineering, and the Session Chairman (2021,2022) of Asian Conference on Artificial Intelligence Technology.  His current research interests include intelligent nuclear security, big data analysis and causal inference, high-reliability computing and security evaluation.
 
\end{IEEEbiography}
\vfill

\end{document}